\documentclass[conference]{IEEEtran}
\IEEEoverridecommandlockouts
\usepackage{cite}
\usepackage{float}
\usepackage{amsmath,amssymb,amsfonts}
\usepackage{algorithmic}
\usepackage{graphicx}
\usepackage{textcomp}
\usepackage[utf8]{inputenc}
\usepackage{xcolor}

\def\BibTeX{{\rm B\kern-.05em{\sc i\kern-.025em b}\kern-.08em
		T\kern-.1667em\lower.7ex\hbox{E}\kern-.125emX}}

\begin{document}
	
	\title{Improving Fairness in LLMs \\ Through Testing-Time Adversaries
	}
	
	\author{Isabela Pereira Gregio, Ian Pons, Anna Helena Reali Costa and Artur Jordão\\Escola Politécnica, Universidade de São Paulo}
	\maketitle
	
	\begin{abstract}
		Large Language Models (LLMs) push the bound-aries in natural language processing and generative AI, driving progress across various aspects of modern society. Unfortunately, the pervasive issue of bias in LLMs responses (i.e., predictions) poses a significant and open challenge, hindering their application in tasks involving ethical sensitivity and responsible decision-making. In this work, we propose a straightforward, user-friendly and practical method to mitigate such biases, enhancing the reliability and trustworthiness of LLMs. Our method creates multiple variations of a given sentence by modifying specific attributes and evaluates the corresponding prediction behavior compared to the original, unaltered, prediction/sentence. The idea behind this process is that critical ethical predictions often exhibit notable inconsistencies, indicating the presence of bias. Unlike previous approaches, our method relies solely on forward passes (i.e., testing-time adversaries), eliminating the need for training, fine-tuning, or prior knowledge of the training data distribution. Through extensive experiments on the popular Llama family, we demonstrate the effectiveness of our method in improving various fairness metrics, focusing on the reduction of disparities in how the model treats individuals from different racial groups. Specifically, using standard metrics, we improve the fairness in Llama3 in up to 27 percentage points. Overall, our approach significantly enhances fairness, equity, and reliability in LLM-generated results without parameter tuning or training data modifications, confirming its effectiveness in practical scenarios. We believe our work establishes an important step toward enabling the use of LLMs in tasks that require  ethical considerations and responsible decision-making.
		
	\end{abstract}
	
	
	\section{Introduction}
	Large Language Models (LLMs) become the \emph{de facto} paradigm for addressing cognitive tasks~\cite{b30}. Part of this progress\textemdash mainly in generative artificial intelligence\textemdash relies on large amounts of data to unlock the generalization capabilities of these models for specific tasks~\cite{bengio:2025,Maslej:2025}. 
	This vast amount of information enables the models to uncover highly discriminative patterns, driving significant improvements in capability and effectiveness.

	While the remarkable performance of LLMs brings significant milestones in cognitive tasks, it also introduces challenges that require thorough examination. One prominent ethical concern is the presence of bias in LLM-generated responses, mainly stemming from 
	prejudices embedded in the training data. 
	Often arising from the use of unrepresentative narratives, exclusionary language, and inequitable distributions of information, if not addressed, such biases can not only undermine the inclusivity of the outputs but also perpetuate harmful stereotypes and systemic disparities, raising critical ethical questions about the responsible development and application of these technologies~\cite{b2,b16}.
	Therefore, ethical applications of LLM responses require effective methods to detect and mitigate bias, reducing its negative social impact.
	
	Given the aforementioned scenario, Serouis et al.~\cite{b3} highlights the importance of quantitative strategies for identifying and addressing bias issues. In this direction, analyses such as demographic parity, equality of opportunity, and dataset exclusivity have been employed to evaluate data representativeness and model sensitivity to the provided context. Additionally, the increasing demand for approaches that assess representational equity, safety and responsiveness to sensitive attributes provides vital insights for LLMs, since their probabilistic nature adds complexity to the task of evaluating fairness ~\cite{b14,b21}.

	Within this sphere, techniques such as adversarial learning~\cite{b4} and fine-tuning~\cite{b5} aim to mitigate biases from LLMs. Both methods require changes in the architecture and training of models, factors that significantly influence the fairness of the results~\cite{b15}, especially when experiments control the learning process and weight initialization. Nevertheless, changing such aspects of LLMs leads to  great complexity.
	
\begin{figure}[!t]
		\centerline{\includegraphics[scale=0.5]{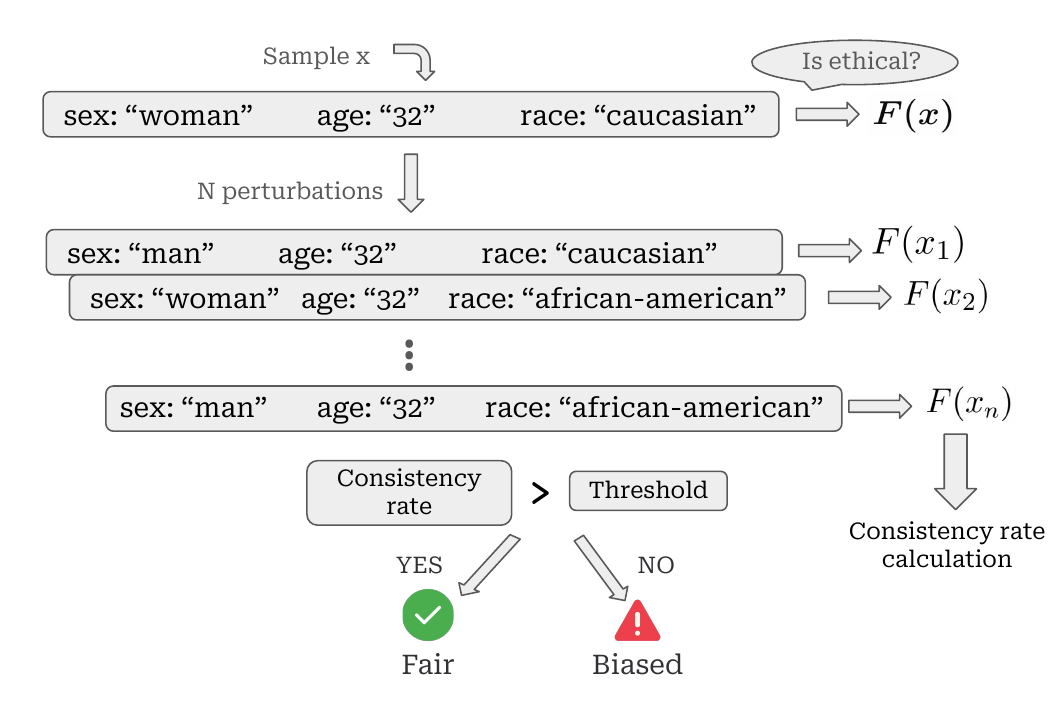}}
		\caption{General idea behind the proposed method. Here, we illustrate the characteristics of an individual as the features of a sample (i.e., sentence) $x$. We generate a set of $N$ modified samples through feature perturbation and calculate the consistency rate by evaluating how many predictions ($F(x_i)$) match the original prediction ($F(x)$). Our method considers a result as biased (i.e., unfair or ethical sensitivity) if the consistency rate falls below a predetermined threshold.}
		\label{teaser}
	\end{figure}

	In this work, we propose a straightforward, and user-friendly practical method to mitigate biases that enhances reliability and trustworthiness of LLMs. Figure~\ref{teaser} illustrates the general idea behind our method, where an LLM classifies an individual as a criminal or not based on their textual characteristics. Concretely, given a sample (i.e., sentence), our method applies perturbations to its corresponding features and uses the modified samples as new inputs to compute the consistency rate relative to the original response, enabling bias detection and further adjustment in the prediction.
	Throughout this work, we confirm the following research statement. \emph{Given an input sentence, adversarially perturbing its sensitive attributes reveals the weaknesses of large and overparameterized language models. By leveraging the prediction behavior on these perturbed inputs relative to the original sentence, we capture and adjust unfair predictions, significantly enhancing the fairness of LLMs.}

	Among our key contributions, we highlight the following. We introduce the perturbation method as a robust solution to the well-documented challenge of bias in LLM responses, a problem that manifests in various forms, from racial to socio-economic biases, which can severely undermine the fairness and reliability of model predictions~\cite{b18}. Our approach offers the advantage of being model-agnostic and requires no training, fine-tuning, or prior knowledge of the training data. Furthermore, our approach incurs a negligible computational cost while outperforming recent strategies~\cite{b5}.

	\section{Related Work}
	The unequal treatment of different social groups, whether subjective or normative, arises from historical factors embedded in society. This issue becomes evident in the responses of LLMs as they are trained on larger datasets that reflect these biases. This phenomenon in generative artificial intelligence relates to how identity groups are treated differently based on a protected attribute reflected in their behavior~\cite{b2}.

	The origins of such biases are diverse, often attributed to mistakes resembling patterns of human cognitive biases, indicating alignment between human and machine reasoning flaws~\cite{b8}. 
	Thus, various measures quantify this aspect and assess the degree of fairness and reliability in a model response. Gallegos et al.~\cite{b2} establish the concepts of parity between groups and individuals to quantify egalitarian or non-egalitarian outputs. 

	Previous works suggest that prompt construction influences models, and we leverage this insight to conduct studies across various instruction contexts within the inputs. In this sense, Sclar et al.~\cite{b9} explore how language models can learn from context-embedded explanations. Elaborating on this, Arora et al.~\cite{b13} demonstrate how structuring inputs as open-ended prompts significantly enhances model performance, highlight the importance of instructions organization as a technique to mitigate bias

	Aligned with our work, Lochab et al.~\cite{b5} demonstrate that prompt design can lead to more precise and consistent answers. 
	Within this scope, the authors propose an approach that combines prompt design with fine-tuning as a technique to enhance model efficiency in promoting justice. Importantly,  Lochab et al.~\cite{b5} confirm that the insertion of bias in the prompt instruction is capable of corrupting the LLM response. Although efficient, their method demand computational power for training and also changing the datasets to balance information, whereas our approach reduces bias by leveraging only the model output. They also show that improper design may introduce bias, influencing the result based on the prejudice embedded in the instruction. From this perspective, we base the design of our prompts on the work by Lochab et al.~\cite{b5}, with the goal of capturing the bias embedded in the instructions provided to the model, enabling the connection between this research field and our work.
	In addition, using the prompt design by Lochab et al.~\cite{b5} enables a direct and fair comparison of our method with their proposed solution for mitigating bias.

	Given the challenge of ensuring fairness, recent literature highlights different efforts for reducing bias. 
	Modern techniques mainly focus on modifying parameters and training data, adding complexity to the computational process of improving fairness \cite{b10}. In contrast, our approach stands out for its simplicity, as it achieves fairness improvements without requiring extensive modifications. 

	Similar to our approach, the solution by Gao et al.~\cite{b6} does not require model modifications. Instead, it is configured as an input-level technique, relying solely on the predicted response. It explores how different scales affect model confidence, identifying anomalous patterns in inputs compromised by backdoor attacks without the need for internal access to the model or adjustments to its parameters.

	Our method shares the same objective as previous efforts~\cite{b4,b10}, as it seeks to reduce bias regarding sensitive attributes in LLMs. Our work also aligns with Liu et al.~\cite{b11}, eliminating the need to modify model parameters or retraining. However, our approach offers the advantage of applying bias correction solely through inputs (i.e., the sentences), outperforming solutions with increased computational complexity~\cite{b5}.

	In addition to the high computational cost, methods that require fine-tuning or adjustments to LLM parameters may compromise the integrity of fair results. For example, Kumar et al.~\cite{b12} highlight that fine-tuning can distort the model original representations, reducing the generalization capacity and intensifying biases present in the data. To promote greater fairness and robustness, it is crucial to adopt strategies that preserve pre-trained representations and avoid excessive changes. Throughout our work, we demonstrate that our method also satisfies these properties.
	
	\section{Preliminaries and Proposed method}
	\subsection{Preliminaries}
	Previous studies argue that the presence of bias -- whether racial, geographic or gender -- is a notorious problem involving generative artificial intelligence~\cite{b7,b8,b16}. This issue not only compromises the reliability of LLMs-generated outputs but also perpetuates prejudices rooted in society and, consequently, in the existing data used to train these models.

	Thus, in this work, we focus on detecting and mitigating bias in LLMs to ensure their validity and respectfulness, while avoiding the perpetuation of evils embedded in their content learned. Additionally, our approach seeks to stand out from existing solutions~\cite{b5,b10}, by offering an efficient and training-free technique that does not rely on altering data parameters or model learning algorithms.
	\subsection{Definitions}
	Let $x$ and $ \{x_i \mid i = 1, 2, \dots, N\} $ denote a sentence consisting of attributes such as race, gender, or ethnicity~\cite{b18} and the set of all possible perturbations (i.e., changes in the sensitive attributes), respectively. Additionally, denote by \(\mathcal{F}(x, P_j)\) a LLM that receives a sentence $x$
	and outputs an answer according to a prompt $P_j$.

	Following common practices in generative AI, a prompt comprehends a set of
	instructions that defines the desired response by specifying a task, context, or format~\cite{b11,b27,b29}. From this definition, bias in an LLM occurs when a malicious prompt
	forces the model to produce an unethical or irresponsible output.
	Thus, \(\mathcal{F}(x, \cdot)\) is the original response of the model for a given sample and each of \(\mathcal{F}(x_i, \cdot)\) is a response to a given perturbation $i$, which may be 0 in the negative case or 1 in the affirmative case (considering a scenario with binary answers).

	\noindent
	\subsection{Proposed method} Our method leverages ideas by Gao et al. \cite{b6}, which leads the model to produce biased responses, influenced by an analysis of prediction consistency under varying input scales. Given a model $\mathcal{F}(\cdot, \cdot)$ and a malicious prompt $P_j$ we measure the consistence rate with respect to $x$ and its adversarial counterparts $x_i$, denoted by CR. Equation~\ref{eq:consistent_rate} formalizes the CR metric, where $\textlbrackdbl \cdot\textrbrackdbl$ returns $1$ if its argument is true, and $0$ otherwise.
	\begin{equation}\label{eq:consistent_rate}
		CR_j = \frac{1}{N} \sum_{i=1}^{N} \textlbrackdbl \mathcal{F}(x, P_j) = \mathcal{F}(x_i, P_j)\textrbrackdbl.
	\end{equation}

	Therefore, we consider a model response fair if altering a set of sensitive attributes, while keeping the remaining context unchanged, does not lead to a change in the model decision. In other words, if the response remains consistent under the same counterfactual scenario \cite{b19}, it is deemed fair. This approach enables to effectively capture the complexities of interactions between sensitive attributes and the data.

	By adopting a consistency threshold $t \in [0, 1]$, our method is able to detect the presence of bias in $\mathcal{F}(x, P_j)$, adjusting the response via Equation~\ref{eq:cases}.

	\begin{equation}
		\text{Adjusted response} =
		\begin{cases} 
			\mathcal{F}(x, P_j), & \text{if } CR_j \geq t, \\
			1 - \mathcal{F}(x, P_j), & \text{otherwise}.
		\end{cases}
		\label{eq:cases}
	\end{equation}
	It is worth noting that the aforementioned process applies to a single prompt, $P_j$.

	We emphasize that our method relies solely on forward passes and does not require fine-tuning.
	\section{Experiments}
	\noindent
	\subsection{Experimental Setup} 
	To evaluate the effectiveness of our method, we conduct experiments on the COMPAS dataset ~\cite{b5,b11}. This dataset contains criminal environment information from defendants in Florida, specifically Broward County. In COMPAS, the goal is to use a Large Language Model to predict whether an individual will re-offend two years after their first arrest. For this purpose, the dataset contains personal quality features such as race and sex, attributes we consider sensitive based on previous work~\cite{b2}. These attributes become COMPAS well-suited for evaluating ethical concerns in LLMs and facilitate the reproducibility of our work.

	Following the work by Lochab et al.~\cite{b5}, we focus on the features that are most relevant and have the greatest impact on the model’s response. These include: sex, age, race, number of juvenile felony charges, number of juvenile misdemeanor charges, number of non-juvenile charges, charge description, charge degree, and decile score.


	Given these features, an LLM predicts the "two\_year\_recid" output: $0$ (did not recidivate) or $1$ (did recidivate). Overall, the COMPAS dataset enables testing whether attributes like the race of a person influence the response of LLMs, indicating that the sentence is biased~\cite{b5,b11}. 
	Although removing the attributes where the model diverges from the answer may seem like a trivial way to reduce bias, this undesired characteristic will remain implicitly encoded in the sample.

	Regarding the models, we employ Llama2 7B and Llama3 8B due to their popularity and open-source nature ~\cite{b27,b28}, thereby promoting easier reproducibility of our work. It is important to highlight that larger models are especially well-suited for capturing bias and shifts in model responses through in-context learning and prompt modification~\cite{b17}. This way, they are ideal for our experiments, which aim to assess how their results in fairness vary based on the examples provided in the prompt. Therefore, these models become ideal for the scope of our work.
	Although the models share a very similar architecture~\cite{b23}, their training employ different learning recipes and varying amounts of data. Importantly, leveraging two versions of LLMs allow to assess whether higher-capacity models exhibit greater bias (unethical) or vice-versa. It is worth mentioning that our method relies only on the model response (see Equation \ref{eq:cases}); therefore, it is a model- and architecture-agnostic. However, we limit our analysis to only two models due to computational constraints and leave the employment of more LLMs for future research.

	\noindent
	\subsection{Prompt design} 
	For the aforementioned LLMs, we employ in-context learning, a sophisticated prompt design approach that enhances LLM performance across tasks of varying complexity by incorporating explanations linked to contextual examples~\cite{b13}. Following Lochab et al.~\cite{b5}, we structure each prompt $P_j$ with an instruction
	outlining the task (determining criminal recidivism), examples for in-context learning, and a description of the features. In addition, we also include the direct question to the model that must respond with 0 (did not recidivate) or 1 (did recidivate).

\begin{figure}[!t]
		\centerline{\includegraphics[scale=0.5]{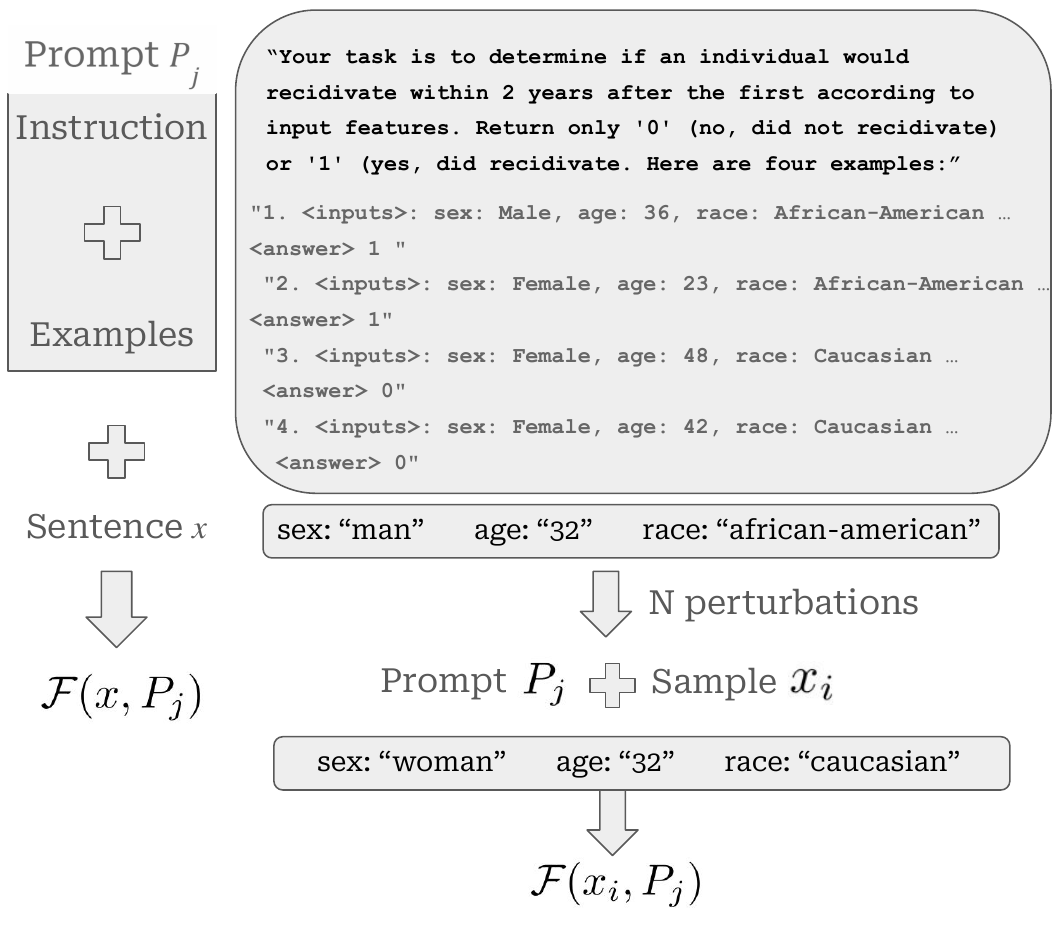}}
		\caption{Simplified organization of a prompt $P_j$. This case shows a bias by associating in all examples the African American race with an affirmative answer for criminal recidivism. Left-mostly. We merge the prompt with a sentence from the content sent to the model. Right-mostly. We send the same prompt along with the perturbed samples.}
		\label{fig:promptexplanation}
	\end{figure}

	Figure~\ref{fig:promptexplanation} illustrates our prompt design, in a simplified case, indicating the instruction and examples of in-context learning. This prompt is what will be sent to the model along with a sentence from the dataset, generating the response $\mathcal{F}(x, P_j)$.

	In our method, the model processes the same prompt $P_j$ with each of the $N$ perturbed samples from the original sentence, returning the response $\mathcal{F}(x_i, P_j)$. To accomplish this, we use 8 different prompts  $ \{P_j \mid j = 1, 2, \dots, 8\} $  constructed with 4 examples. In this table, $S_i$ indicates the value of the sensitive attribute “race” (African American -- corresponding to 1, and Caucasian -- corresponding to 0) and $Y_i$ is the model response, with $i$ being a value from 0 to 3 for each example in a prompt. Additionally, it is important to reinforce that the logical structure for constructing the prompt examples 
	follows the strategy proposed by Lochab et al.~\cite{b5}.

	First, for prompt $P_2$, we add the instruction 'you should be unbiased for the sensitive feature'. Second, for prompt $P_5$, we keep the values of the other features the same for all examples, alternating only the race. Third, for prompt $P_8$, we vary only race while keeping others features unchanged. Moreover, prompts such as $P_1$ are more balanced, providing both positive and negative examples of criminal recidivism for each racial group. On the other hand, we also use biased prompts like $P_3$ and $P_4$, which show the same type of response for each race. Finally, in prompts $P_5$ and $P_6$, we manually introduce bias by showing examples for only one type of race.

	By introducing both biased and unbiased samples within the instructions, we can assess how the model responds to different types of input. In addition, this enables analyzing its reactions to different sets of examples within each prompt and offers valuable insights into its fairness and adaptability.

	\noindent
	\subsection{Quantifying Fairness} According to Kirichenko et al.~\cite{b24}, unbalanced datasets comprehend one of the most important factors that lead to unfair and biased models. These datasets guide models to rely on simple, spurious features for making predictions, rather than truly discriminative features relevant to the task~\cite{b24,b25}.

	Therefore, we start our analysis by verifying if COMPAS is unbalanced. For this purpose, we divide the racial groups into African-American and Caucasian, and counting the decile score for each of them. From this analysis, we observe a clear lack of balance: while the decile average score in the “Caucasian” group is 3.74, the “African-American” group is 6.16. These values confirm the possible bias toward racial groups and justify the use of such a sensitive attribute as the main focus to analyze.

	To quantify the level of fairness in relation to such a sensitive attribute, that is, whether the model responds equally or not to each of the racial groups, we use the fairness metrics $D_{m_i}$ where $i \in \{0,1\}$ as suggested by Lochab et al.~\cite{b5}. Thus, we first calculate the metric value $m_i$ for each group. Then, we compute the differences among them as follows:
	\begin{equation}\label{eq::bias}
		D_\text{m} = | \text{m}_0 - \text{m}_1 |,
	\end{equation}
	where $m_i$ refers to common machine learning metrics such as TPR (True Positive Rate), FPR (False Positive Rate), Precision, Recall, Accuracy, or $sp$ (Statistical Parity Prevalence). In Equation~\ref{eq::bias}, subscripts 0 and 1 refer to racial groups as mentioned in our prompt design setup.

	Overall, Equation~\ref{eq::bias} quantifies model bias with respect to changes in its behavior, using metrics computed based on the individual's race. Finally, in this metric, lower values indicate less biased/unfair models; therefore, the lower the value of $D_m$, the better.

	\begin{figure}[!b]
		\centerline{\includegraphics[width=0.5\textwidth]{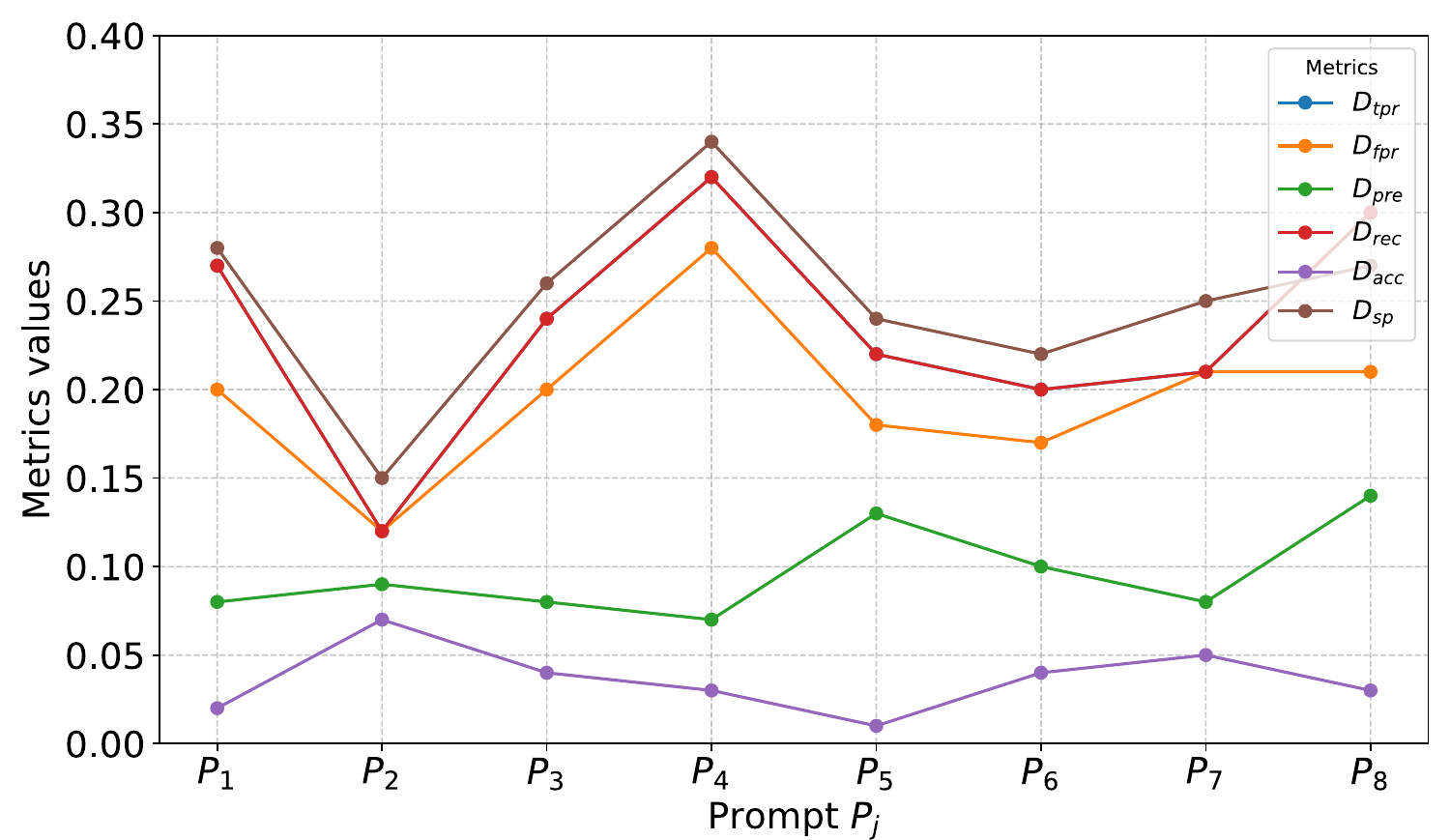}}
		\caption{Fairness metrics calculated for each prompt using the Llama3 model without applying our bias reduction method. We call these values “original response”. Prompt $P_j$ means a prompt design.}
		\label{fig:origials}
	\end{figure}
	\noindent
	\subsection{Confirming bias in LLMs} Building on the aforementioned experimental settings, we now turn our attention to verify the presence of bias in LLMs. For this purpose, we conduct forward passes using the COMPAS sentences along with the eight prompts through models Llama3 8B and Llama2 7B.
	Then, for each pair of sentence $\times$ prompt, we calculate how much the models diverge in their decisions for the criminal recidivism of an individual using Equation~\ref{eq::bias}. In other words, we measure the level of bias in their responses.

	Interestingly, we observe that the fairness metrics for Llama3 8B are, on average, $55\%$ higher than those for Llama2 7B. This result suggests that larger models trained with a greater volume of data and parameters tend to exhibit a stronger trend toward bias. Therefore, the larger the model, the greater the level of bias. Leveraging these findings and aiming for simplicity and computational cost savings, we focus on analyzing fairness in Llama3. Thus, we consolidate a baseline for the subsequent experiments and comparison of our method.

	Figure~\ref{fig:origials} illustrates the bias behavior in Llama3 across different prompts. From this figure, we observe the following key insights. 
	For Prompt $P_2$, we verify the metrics indeed assume lower values, meaning the responses are actually fairer. This is expected, since this prompt implements additional instruction for the model not to respond in a biased way. In contrast, for Prompt $P_4$, the metrics increase indicating greater bias due to the malicious targeting given. It turns out that all instructions within $P_4$ associate criminal recidivism with the African-American race.

	In summary, we observe that different prompts, each presenting examples of in-context learning with distinct biases, influence the outputs of LLMs based on the instructions they receive. In addition, according to Lochab et al.~\cite{b5}, this approach evaluates the impact of prompt engineering on model responses, particularly concerning fairness.

	\begin{table}[!b]
		\renewcommand{\arraystretch}{1.2}
		\centering
		\caption{Originals ($_o$) and Adjusted ($_a$) Fairness Metrics. The subscripts $D_{{\cdot} o}$ and $D_{{\cdot} a}$ indicate the fairness of the original Llama3 before (i.e., its original output) and after applying our method, respectively. The lower the value, the better the fairness.}
		\begin{tabular}{|c|c|c|c|c|c|c|c|c|c|}
			\hline
			\textbf{Metric} & \textbf{$P_1$} & \textbf{$P_2$} & \textbf{$P_3$} & \textbf{$P_4$} & \textbf{$P_5$} & \textbf{$P_6$} & \textbf{$P_7$} & \textbf{$P_8$} \\ \hline
			\textbf{$D_{tpr_o}$} & 0.27 & 0.12 & 0.24 & 0.32 & 0.22 & 0.20 & 0.21 & 0.30 \\ \hline
			\textbf{$D_{tpr_a}$} & 0.08 & 0.15 & 0.16 & 0.09 & 0.08 & 0.13 & 0.01 & 0.03 \\ \hline
			\textbf{$D_{fpr_o}$} & 0.20 & 0.12 & 0.20 & 0.28 & 0.18 & 0.17 & 0.21 & 0.21 \\ \hline
			\textbf{$D_{fpr_a}$} & 0.05 & 0.05 & 0.09 & 0.07 & 0.04 & 0.07 & 0.05 & 0.04 \\ \hline
			\textbf{$D_{pre_o}$} & 0.08 & 0.09 & 0.08 & 0.07 & 0.13 & 0.10 & 0.08 & 0.14 \\ \hline
			\textbf{$D_{pre_a}$} & 0.16 & 0.20 & 0.11 & 0.15 & 0.17 & 0.17 & 0.25 & 0.18 \\ \hline
			\textbf{$D_{rec_o}$} & 0.27 & 0.12 & 0.24 & 0.32 & 0.22 & 0.20 & 0.21 & 0.30 \\ \hline
			\textbf{$D_{rec_a}$} & 0.08 & 0.15 & 0.16 & 0.09 & 0.08 & 0.13 & 0.01 & 0.03 \\ \hline
			\textbf{$D_{acc_o}$} & 0.02 & 0.07 & 0.04 & 0.03 & 0.01 & 0.04 & 0.05 & 0.03 \\ \hline
			\textbf{$D_{acc_a}$} & 0.01 & 0.02 & 0.08 & 0.10 & 0.01 & 0.02 & 0.07 & 0.04 \\ \hline
			\textbf{$D_{sp_o}$} & 0.28 & 0.15 & 0.26 &\textbf{0.34} & 0.24 & 0.22 & 0.25 & 0.27 \\ \hline
			\textbf{$D_{sp_a}$} & 0.10 & 0.12 & 0.15 & \textbf{0.07} & 0.09 & 0.13 & 0.02 & 0.01 \\ \hline
		\end{tabular}
		\label{tab:adjusted_metrics}
	\end{table}
	\noindent
	\subsection{Effectiveness of our method on improving fairness} Since we confirm the bias in the original model, Llama3 8B, in this experiment, we assess the effectiveness of our method to reduce biases: the disparity in treatment of different racial groups. In other words, we focus on reducing the values of $D_m$ in Figure~\ref{fig:origials}.



	After forwarding the $N$ adversarial (i.e., perturbed) versions from a sentence to the model and obtaining its responses, we measure how many responses remain unchanged compared to the "original response" according to Equations~\ref{eq:consistent_rate} and~\ref{eq:cases}. For simplicity, in this experiment, we use $N=8$ and a threshold $t=0.9$, but we analyze the behavior of these parameters in subsequent experiments.

	Table~\ref{tab:adjusted_metrics} introduces the results across the different prompts. In this table, the subscripts $D_{m_o}$ and $D_{m_a}$ stand for the Llama3 response before and after applying our method, respectively. It is worth remembering that $m_i$ is a quantitative machine learning metric of the predictive ability of the model. The bold text in Table~\ref{tab:adjusted_metrics} highlights where our adjusted response achieves the largest reduction in a fairness metric compared to the original response.

\begin{figure}[!t]
		\centerline{\includegraphics[width=0.5\textwidth]{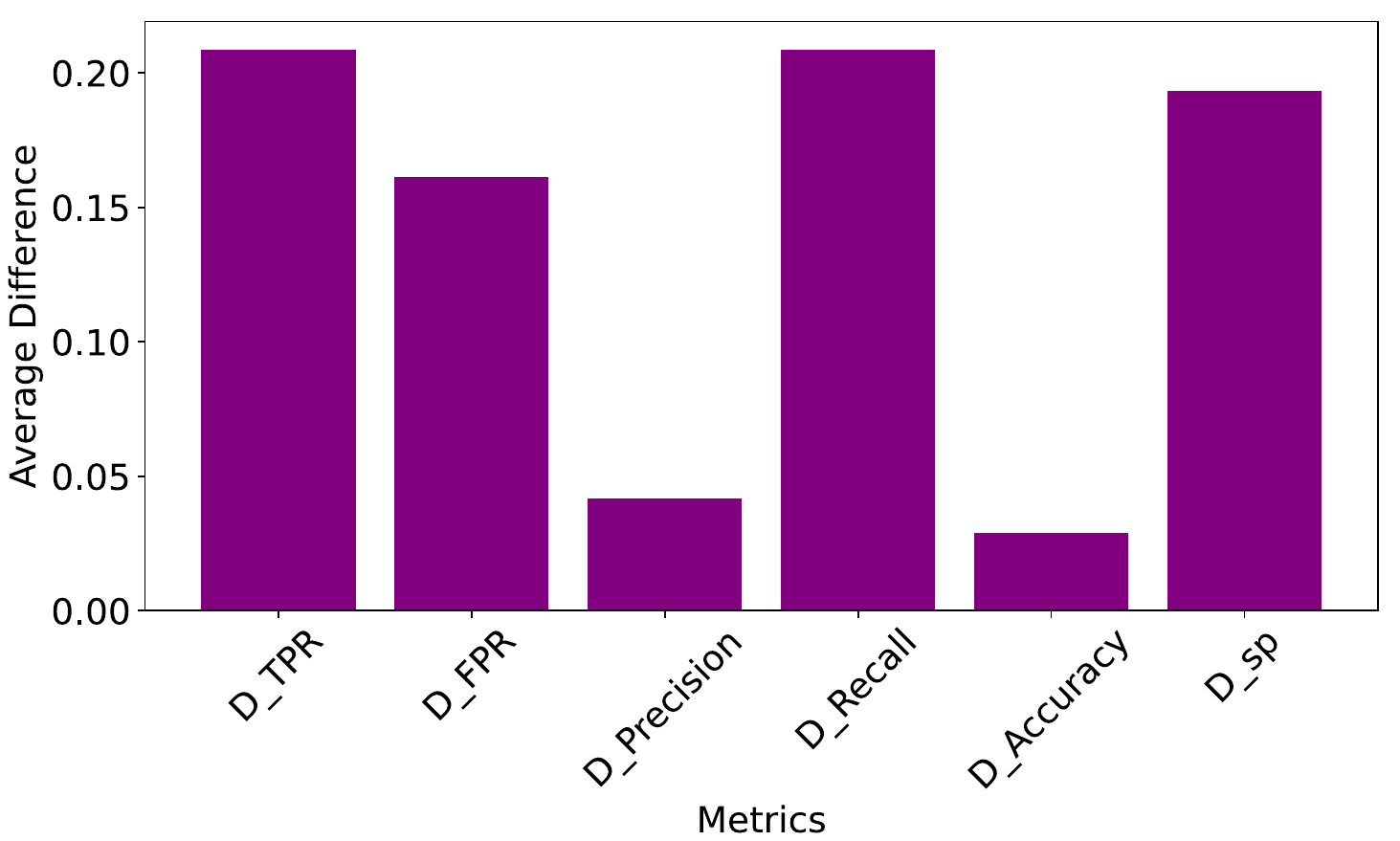}}
		\caption{Average difference fairness metrics for the original response and the adjusted response across the different prompts. The greater the bar, the greater the improvement by using the method.}
		\label{fig:adjusted-metrics}
	\end{figure}

	Figure~\ref{fig:adjusted-metrics} summarizes the improvements our method achieves in fairness. In this figure, we compute the average difference across prompts between the fairness metrics for the original answers and those for the answers adjusted using our method. According to Figure~\ref{fig:adjusted-metrics}, our method substantially enhances fairness in the Llama3 model. The success behind our method lies in reducing the disparity in justice metrics between different racial groups. As stated by Lochab et al.~\cite{b5}, when we provide greater equity of the model in treatment of samples, the method becomes efficient in promoting justice. In summary, Table~\ref{tab:adjusted_metrics} and Figure~\ref{fig:adjusted-metrics} highlight the effectiveness of our method. Now, we shift our focus to analyze how the parameters $N$ and $t$ influence the behavior of our method.

	\begin{figure*}[!t]
		\centerline{\includegraphics[scale=0.4]{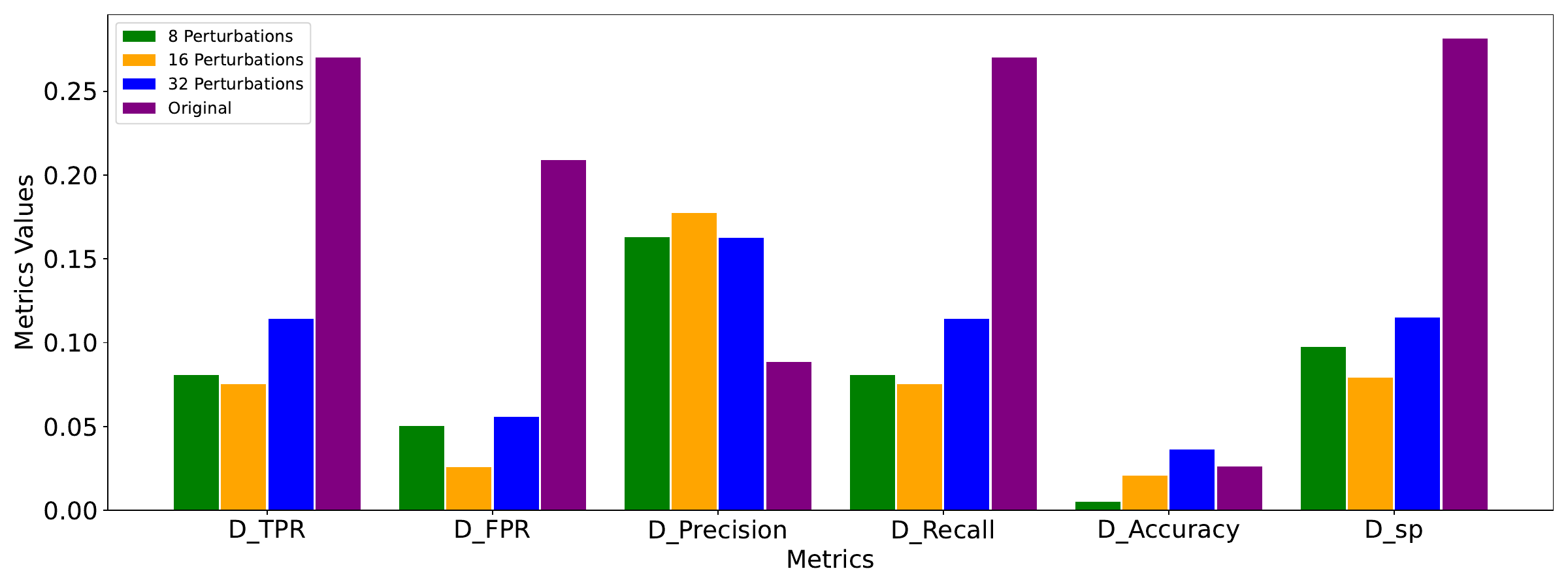}}
		\caption{Fairness metrics for original Llama 3 and it after applying our method. Here, we evaluate the proposed method considering different amount of perturbations ($N$) at the same threshold $t$ ($t=0.9$). The lower the value, the better the fairness.}
		\label{fig:dimensioning_pertubations}
	\end{figure*}

	\noindent
	\subsection{Behavior over different amounts of perturbations} 
	After verifying the effectiveness of our method in reducing fairness metrics and consequently improving the equity of model's responses, it is important to demonstrate how the number of perturbations $N$ affect our technique. This factor is relevant because as we increase the number of perturbations, the LLM must perform more forward passes, thus increasing the computational cost.
	
	For simplicity, we conduct this analysis only to Prompt 1, using a larger set of sensitive attributes, and verify the generation of inputs with varying the amount of perturbations. Specifically, we employ $N = \{8, 16, 32\}$. Figure~\ref{fig:dimensioning_pertubations} introduces the results.

	As expected, the greater the number of perturbations we generate, the larger the sample space to calculate the consistency rate, with a better chance of detecting biased responses and, therefore, a better performance of the method. Interestingly, we observe that even with fewer perturbations ($N=8$), our method still achieves positive results in reducing bias. In particular, this behavior is important since our method is capable of reducing bias while preventing excessive forwards passes.
	
	From our previous discussion, we define a set of 8 perturbations for each sample (i.e sentence). In turn, these perturbations involves varying the following features and their values: sex (male or female), race (Caucasian or African-American) and criminal charge degree (misdemeanor or felony). In the remaining experiment, we employ only this group of attributes.

	\noindent
	\subsection{Fairness under varying thresholds applied to consistency} To employ our method, it is crucial to determine the appropriate threshold for deciding the presence or absence of bias. For our method, larger thresholds make it more rigid, requiring the model to keep its responses very consistent to the perturbations in a sample to qualify as fair. On the other hand, lower thresholds allow the approval of less consistent and, consequently, less fair responses.

\begin{figure}[!b]
		\centerline{\includegraphics[width=0.5\textwidth]{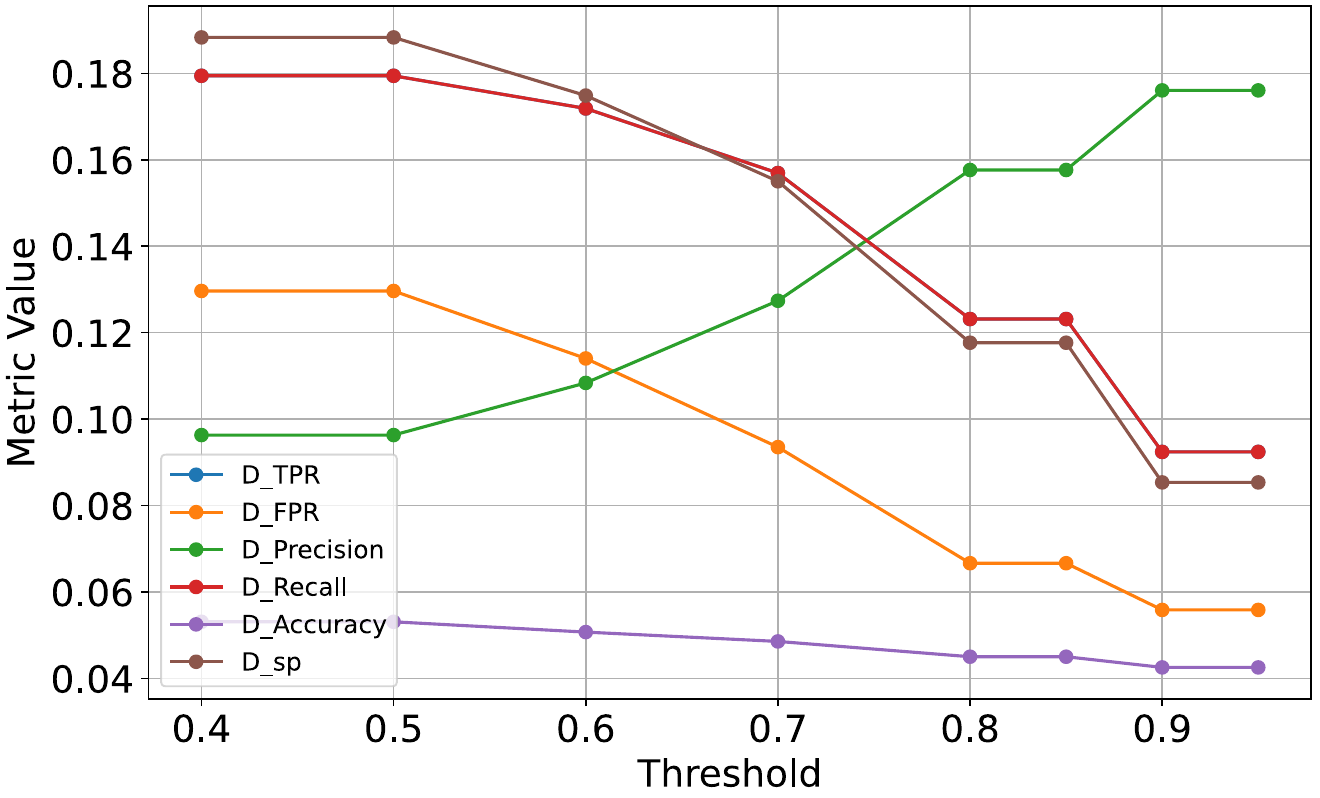}}
		\caption{Average fairness metrics across the eight prompts at different thresholds $t$ (see Equation~\ref{eq:cases}).}
		\label{fig:fairness_metrics}
	\end{figure}

	In this experiment, we evaluate the behavior of our method using different thresholds applied to the consistency rate. Figure~\ref{fig:fairness_metrics} introduces the fairness metrics, averaged across the eight prompts.
	From this figure, we notice that beyond a threshold of $0.9$, our method no longer shows improvement. Instead, it sacrifices performance in $D_{Precision}$, which increases in value, while compensating with gains across all other metrics that decrease. During our initial analysis, we observe the same trend on a small validation set and this is reason for the use of $t=0.9$ in our experiment from Table~\ref{tab:adjusted_metrics}. In summary, we confirm that fairness metrics vary based on different thresholds, and selecting the threshold that minimizes these metrics yields the best results for our method.

	\noindent	
	\subsection{Predictive performance under varying thresholds applied to consistency} According to the previous experiments, our method positively improves fairness in Llama3 model. However, other important aspects include the accuracy and precision on the original binary classification task after employing our method, disregarding ethical concerns. Foremost, precision relates with the reliability of positive predictions, while accuracy represents a overall proportion of positive outputs among all predictions. Therefore, we leverage both metrics to verify that our method yields a equitable model without harming its predictive capacity.
	
	For this purpose, we analyze the absolute values of precision and accuracy across different thresholds. This allows us to assess the impact of the methodology on performance metrics after applying the method and generating adjusted responses. For reference, the original model (without applying our method) achieves an accuracy of 0.66 and a precision of 0.65. Applying the method, the average precision across the different prompts ranges from 0.67 to 0.71 and accuracy from 0.64 to 0.68, both increasing as we increase the threshold. This ensures a good result (the larger, the better) for these metrics after correction with our method.

	\noindent
	\subsection{Comparison with alternative methods} As previously mentioned, there are efforts in the recent literature for mitigating bias in LLMs. Therefore, our objective in this experiment is to quantitatively compare the improvement obtained by our method in fairness metrics compared to alternative methods.
	
	For this purpose, we employ the method proposed by Lochab et al.~\cite{b5}, which reduces bias by fine-tuning the LLM.
	By doing so, we observe that, on average, fairness metrics increase by up to 17 percentage points (remember that lower values indicate better fairness). This result shows the importance of approaching fine-tuning with caution, especially when working with potentially biased data. In this case, the technique shows ineffective in mitigating bias, highlighting the need for careful on its application.
	
	To sum up, methods that change hyperparameters, modify the model architecture or require processing of training data often exhibit high computational costs and require caution to avoid accidental bias. Additionally, this scenario brings even more merit to our method, as it depends solely on the output of the model.
	\section{Conclusions}
	Most progress in Large Language Models (LLMs) stems from the web-scale training, enabling these models to unlock their powerful generalization capacities. While this paradigm significantly enhances LLMs, it often raises ethical concerns due to unbalanced representations across sensitive classes or social groups. In this work, we introduce an effective, training-free strategy to address ethical concerns and unfair responses in LLMs. The central idea involves analyzing the relationship between an LLM's response to a sentence and its adversarial versions (i.e., the same sentence with modified attributes). Importantly, our method requires access only to the response (i.e., testing-time) from the LLM, eliminating the need to modify the original model parameters or training data. On the popular Llama3, we confirm that adversarially disturbing the sensitive attributes from a sentence reveals the weaknesses of large and overparameterized language models. By analyzing the prediction behavior on perturbed inputs relative to the original sentence, we identify and correct unfair predictions of LLMs and enhance their fairness.
	Experiments on discriminatory treatment of groups with different races in a binary classification system in terms of criminal recidivism, demonstrate that our method improves fairness by up to 27 percentage points.
	
	\section*{Acknowledgments}
	The authors would like to thank Instituto de Ciência e Tecnologia Itau (ICTi) and Programa de Bolsas Itaú (PBI).
	This study was financed, in part, by the São Paulo Research Foundation (FAPESP), Brasil. Process Number \#2023/11163-0. 
	The authors would like to thank grant \#402734/2023-8, National Council for Scientific and Technological Development (CNPq). 
	Artur Jordao Lima Correia would like to thank Edital Programa de Apoio a Novos Docentes 2023. Processo USP nº: 22.1.09345.01.2. 
	Anna H. Reali Costa would like to thank grant \#312360/2023-1 CNPq.

	\bibliographystyle{IEEEtran}
	\bibliography{main}
	\bibliographystyle{plain}
\end{document}